\documentclass[11pt]{article}

\usepackage{acl}

\usepackage{times}
\usepackage{latexsym}

\usepackage[utf8]{inputenc} 
\usepackage[T1]{fontenc}    
\usepackage{hyperref}       
\usepackage{url}            
\usepackage{booktabs}       
\usepackage{amsfonts}       
\usepackage{nicefrac}       
\usepackage{microtype}      
\usepackage{graphicx}
\usepackage{amsmath}
\usepackage{appendix}
\usepackage{scrextend}
\usepackage{soul}
\usepackage{listings}

\newcommand{\extra}{\textsc{extra}}

\title{Retrieval-augmented Image Captioning}

\author{Rita Ramos$^{\dagger}$ \ \ Desmond Elliott$^{\star}$ \ \ Bruno Martins$^{\dagger}$ \\
        $^{\dagger}$INESC-ID, Instituto Superior Técnico, University of Lisbon \\ $^{\star}$Department of Computer Science, University of Copenhagen \\
        \texttt{ritaparadaramos@tecnico.ulisboa.pt}}

\begin{document}

\maketitle

\begin{abstract} 

Inspired by retrieval-augmented language generation and pretrained Vision and Language (V\&L) encoders, we present a new approach to image captioning that generates sentences given the input image and a set of captions retrieved from a datastore, as opposed to the image alone. The encoder in our model jointly processes the image and retrieved captions using a pretrained V\&L BERT, while the decoder attends to the multimodal encoder representations, benefiting from the extra textual evidence from the retrieved captions. Experimental results on the COCO dataset show that image captioning can be effectively formulated from this new perspective. Our model, named \extra{}, benefits from using captions retrieved from the training dataset, and it can also benefit from using an external dataset without the need for retraining. Ablation studies show that retrieving a sufficient number of captions (e.g., k=5)
can improve captioning quality. Our work contributes towards using pretrained V\&L encoders for generative tasks, instead of standard classification tasks.

\end{abstract}

\section{Introduction}

Image captioning is the task of automatically generating a short textual description for a given image. 
The standard approach involves the use of encoder-decoder neural models, combining a visual encoder with a language generation decoder (see \citet{hossain2019comprehensive} for a survey). In early studies, the encoder was typically a Convolutional Neural Network model (CNN) pretrained on the ImageNet classification dataset \cite{russakovsky2015imagenet} or a pretrained Faster-RCNN object detector \cite{ren2015faster}, whereas the decoder was commonly an LSTM \cite{hochreiter1997long} together with an attention mechanism \cite{bahdanau2014neural}. 
More recently, Transformer based models have been achieving state-of-the-art results on a variety of language processing \cite{vaswani2017attention, devlin2018bert, radford2019language} and computer vision tasks \cite{dosovitskiy2020image}. Accordingly, state-of-the art image captioning models have replaced the conventional CNN-LSTM approach with encoder-decoder Transformers \cite{liu2021cptr}. Still, in both cases, the encoder only attains 
visual representations, whereas richer features could be captured from image--text interactions if the 
encoder had access to useful textual context related to the input image (e.g., sentences associated to similar images). 

In this paper, we present a new type of image captioning model that uses a pretrained V\&L BERT \cite[\textit{inter-alia}]{tan2019lxmert,li2019visualbert,bugliarello2020multimodal} to encode both the input image and captions retrieved from similar images. This model generates captions conditioned on representations that consider linguistic information beyond the image alone. Moreover, specifically using the retrieved captions as textual contexts rather than other alternatives (e.g., image tags or object names) can aid guiding the language generation process, since the model is now provided with well-formed sentences that are semantically similar to what the predicted caption should resemble.

In experiments on the COCO dataset \cite{chen2015microsoft}, the proposed model is competitive against state of the art methods. 
In a series of ablation experiments, we find that the 
model improves when encoding multiple retrieved captions, and that it could reach better performance if it was able to retrieve better captions from the datastore. In experiments on the smaller Flickr30K dataset, we show that allowing the model to retrieve captions from the larger COCO dataset can improve performance without needing to retrain the model.

We hope that our work inspires the adoption of pretrained V\&L encoders for a broader range of generative multimodal tasks. There have been several recent studies proposing V\&L BERTs to learn generic multi-modal representations with large amounts of paired image and text data, which can then be fine-tuned to downstream tasks. However, these  pretrained models have mostly been applied to classification tasks 
and have seen limited use for image captioning, a task which typically only considers single-input images, as opposed to image-text pairs, as proposed in this work.

\section{Model}

We present a model that captions images, given both the image and a set of $k$ captions retrieved from similar images using a retrieval system. 
%
This approach belongs to the class of retrieval-augmented language generation models \cite{weston2018retrieve,izacard2020leveraging}. In our model, the image and the retrieved captions are jointly encoded using a pretrained V\&L encoder to capture cross-modal representations in the combined input data. We denote our model as \extra{}: \textbf{E}ncoder with \textbf{C}ross-modal representations \textbf{T}hrough \textbf{R}etrieval \textbf{A}ugmentation. It consists of three components, namely an encoder, a retrieval system, and a decoder.

\begin{figure*}[t]
  \includegraphics[width=0.95\linewidth]{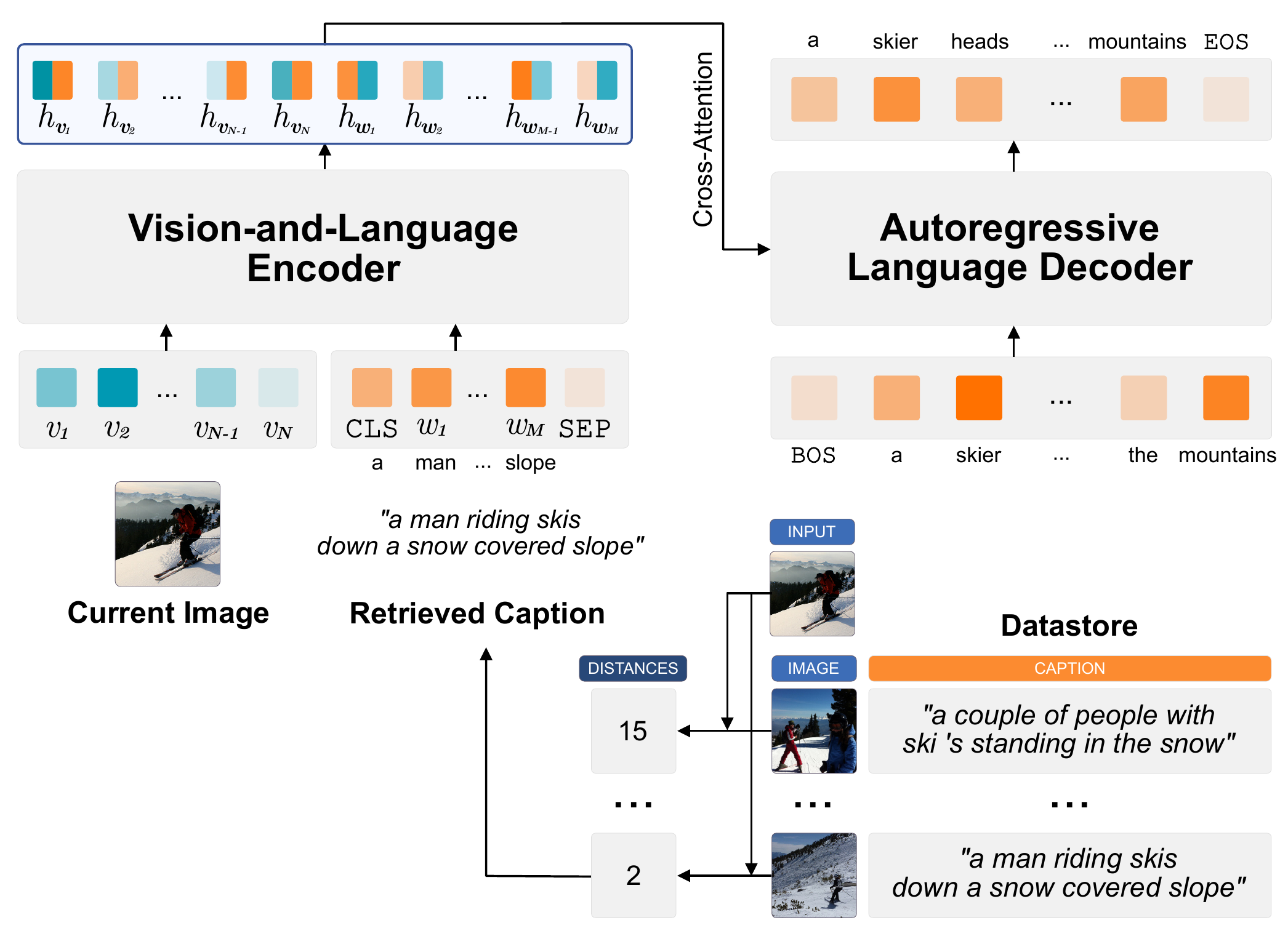}
  \caption{Illustration of the \extra{} model. Given an input image, \extra{} retrieves captions from a datastore and encodes both the input image and the retrieved captions using a pretrained vision-and-language encoder. The decoder attends over both the visual and linguistic outputs, improving the quality of the generated caption.} 
  \label{approach}
\end{figure*}

\subsection{Encoder}\label{enc}

The encoder in \extra{} is LXMERT\footnote{The exploration of other encoders is left for future work.} \cite{tan2019lxmert}, a pretrained vision-and-language Transformer that jointly encodes a visual input \textbf{V} and a linguistic input \textbf{L}. The visual input is represented as $N$=36 regions-of-interest V=$\{v_1, ... , v_N\}$ extracted from the image using the Faster-RCNN object detector, pretrained \cite{anderson2018bottom} on the Visual Genome dataset \cite{krishna2016visual}. 
A sentence in the linguistic input is tokenized into $M$ sub-words using the BERT tokenizer \cite{devlin2018bert}, starting with a special classification token 
\texttt{CLS} and ending with a special
delimiter token \texttt{SEP}. We extended LXMERT to encode $k$ sentences by concatenating the tokenized sentences into a single input, each separated by the delimiter token:
L=$\{\texttt{CLS}, w_{1}^{L_1} $\dots$ w_{M}^{L_1}, \texttt{SEP} $\dots$ w_{1}^{L_k} $\dots$ w_{M}^{L_k}, \texttt{SEP}\}$. The sentences are obtained from a datastore via a retrieval system, as explained in Section~\ref{retrieval}.

The encoder produces a sequence of cross-modal representations of image 
and the text, 
which are the inputs to the decoder, described in Section~\ref{dec}.

\subsection{Image--Text Retrieval and Datastore}\label{retrieval}

The retrieval system builds on the Facebook AI Similarity Search (FAISS) nearest-neighbour search library \cite{JDH17}. FAISS allows for the indexing of high-dimensional vectors, i.e., a datastore \textbf{D}, and it offers the ability to quickly search through the datastore given a similarity measure \textbf{S}, e.g., Euclidean distance or cosine similarity. 


Given an input image \textbf{V}, the retrieval system finds L, the set of $k$ captions retrieved from the datastore, which \extra{} encodes together with the image. 
The datastore consists of captions associated with images in a dataset\footnote{This can either be the training set or an external dataset.}. Each caption in the datastore, and the query input image, are represented using vectors extracted from CLIP \cite{radford2021learning}, allowing image--text search by projecting images and text to a shared latent space. Using FAISS, the input image can then be compared against the vectors\footnote{This comparison can be pre-computed for efficiency.} from \textbf{D} to search over the corresponding $k$ nearest-neighbours captions.

\subsection{Decoder}\label{dec}
The decoder is a conditional auto-regressive language model based on GPT-2 \cite{radford2019language} with additional cross-attention layers to the encoder. The Transformer layers in the decoder already contain a masked multi-head self-attention sublayer, which self-attends to the previous words. We add cross-attention layers \cite{vaswani2017attention} 
subsequent to the masked self-attention sublayers, so the decoder can attend to the encoder outputs. 

The decoder predicts a caption $y_1\dots y_M$ token-by-token, conditioned on the previous tokens and the outputs of the 
V\&L encoder. The model's parameters $\theta$ are trained by minimizing the sum of the negative log-likelihood of predicting the ground truth token at each time-step, using the standard cross-entropy loss:
\begin{align}\label{eqatt1}
L_{\theta}=-\sum_{i=1}^{M}\log P_{\theta}(y_{i}| y_{<i}, \textbf{V}, \textbf{L}).
\end{align}

We can also fine-tune the model with Self-Critical Sequence Training \cite{rennie2017self}.

\section{Experimental Protocol}

\begin{table*}
\begin{tabular}{lccc@{\hspace{4ex}}ccccc}
\multicolumn{1}{c}{} & \multicolumn{4}{c}{Cross-Entropy Optimization} & \multicolumn{4}{c}{CIDEr Optimization} \\
\cmidrule(lr){2-5} \cmidrule(lr){6-9}

               & B4  & METEOR & CIDEr   & SPICE  & B4   & METEOR  & CIDEr   & SPICE      \\
\midrule
\multicolumn{4}{l}{\textit{Encoder-Decoder models}}\\
\midrule
Up-Down              &36.2 & 27.0 & 113.5 & 20.3& 36.3  & 27.7  & 120.1 & 21.4 \\

CaMEL$_{\text{Faster R-CNN}}$ & 36.1 & 28.0  & 114.8 & 20.8 &- & -  &- &-  \\

GCN-LSTM             & 36.8 &27.9&116.3& 20.9   & 38.2  & 28.5 & 127.6 & 22.0  \\

  VL-T5 & 34.5 & 28.7  & 116.5 & 21.9  & - & -& -& -  \\
  
  AoANet              &37.2 & 28.4  & 119.8 & 21.3  & 38.9  & 29.2 & 129.8 & 22.4  \\

   CPTR  & - & -& -& -   & 40.0 & 29.1& 129.4 &- \\
   
\textbf{\extra{} ($k=5$)} & 37.5 & 28.5  & 120.9 & 21.7  & 36.4 & 28.2& 131.1 &21.3 \\

    CaMEL$_{\text{CLIP-RN50$\times$16}}$ &38.8 &29.4  & 125.0 & 22.2 &41.3 & 30.2  & 140.6 & 23.9   \\
    \midrule

\textit{V\&L BERT models} \\
\midrule
  VLP & 36.5 & 28.4 & 116.9 & 21.2 &39.5 & 29.3 & 129.3& 23.2 \\
  
   OSCAR$_{B}$ & 36.5 & 30.3 & 123.7 & 23.1 &40.5 & 29.7 & 137.6& 22.8 \\
  
  VinVL$_{B}$ &  38.2 & 30.3 & 129.3 & 23.6 & 40.9 & 30.9& 140.4 & 25.1 \\
 
 
\bottomrule

\end{tabular}
 \caption{Results on the Karpathy COCO test split. \textbf{\extra{}} ($k=5$) is competitive against encoder-decoder models. We present results with cross-entropy training and after Self-Critical Sequence Training using the CIDEr metric.}\label{SAT}
\end{table*}

\subsection{Datasets and Metrics}
We evaluate our model on the COCO dataset \cite{chen2015microsoft}, using the standard Karpathy splits of 113287 images for training, 5000 for validation, and 5000 for testing, with 5 captions per image. 

Standard metrics were used to evaluate caption generation, namely BLEU-4 (B4) \cite{papineni2002bleu}, METEOR \cite{denkowski2014meteor}, CIDEr \cite{vedantam2015cider}, and SPICE \cite{anderson2016spice}, using the MS COCO caption evaluation package\footnote{\url{https://github.com/tylin/coco-caption}}.

\subsection{Implementation and Training Details}
The implementation\footnote{\url{https://github.com/RitaRamo/extra}} of \extra{} uses the HuggingFace Transformers library \cite{wolf2020transformers}. The encoder is LXMERT \cite{tan2019lxmert}, a 14-layer V\&L model pretrained on 9 million image--sentence pairs across a variety of datasets and tasks. Following \citet{liu2021cptr}, the decoder is a 4-layer randomly initialized GPT-2-style Transformer network with 12 attention heads and additional cross-attention layers. The retrieval systems uses FAISS with a flat index (\emph{IndexFlatIP}) without any training. The corresponding datastore \textbf{D} consists of all the captions associated to the 113287 images in the COCO training set. For caption retrieval, the captions in the datastore and the input image (i.e., the query) are both represented with features extracted from the CLIP-ResNet50$\times$4 pretrained model. Using the cosine similarity for comparison, a total of $k=5$ captions are retrieved to be jointly encoded with the input image by \extra{}. Notice that CLIP-ResNet50$\times$4 features are only used for retrieval, while the \extra{} encoder, i.e. the pretrained LXMERT, requires Faster-RCNN features, and thus it cannot use CLIP visual features. 

\extra{} is trained in two stages using a single NVIDIA V100S 32GB GPU. In the first stage, \extra{} is trained end-to-end with the  cross-entropy loss, using a batch size of 64 and the AdamW optimizer \cite{loshchilov2017decoupled} with a learning rate of $3e-5$. The encoder is trained with a linear warmup for the first epoch to prevent gradients from the randomly initialized decoder from harming the pretrained encoder. The model was trained with early stopping: training ends if there is no improvement after 5 consecutive epochs on the validation set over the BLEU-4 metric. In the second stage, \extra{} is fine-tuned with Self-Critical Sequence Training \cite{rennie2017self} with CIDEr optimization and greedy search decoding as a baseline, using a batch size of 55, a learning rate of 3e-5, and a frozen encoder. Captions are decoded using beam search with a beam size of 3.

\section{Results}\label{results}

Table \ref{SAT} shows the performance of \extra{} compared to strong encoder-decoder models. We compare against the widely-used Up-Down \cite{anderson2018bottom} and AoANet models \cite{huang2019attention}, both using a Faster-RCNN image encoder; the GCN-LSTM model \cite{yao2018exploring} with a Graph Convolutional Network (GCN) encoder; the CPTR model \cite{liu2021cptr} employing a ViT Transformer encoder \cite{dosovitskiy2020image}; the VL-T5 Transformer model \cite{cho2021unifying} with a vision and language encoder; 
and the recent CaMEL model \cite{barraco2022camel} with the CLIP-RN50×16 encoder. 
Our model is also compared with state-of-art models that do not use the encoder-decoder paradigm but instead unify the Transformer encoder and decoder into a single model, namely the VLP \cite{zhou2020unified}, OSCAR-base \cite{li2020oscar}, and the VinVL-base \cite{zhang2021vinvl} models. 
We note that these are general purpose V\&L models, not specifically designed for image captioning.

Overall, \extra{} is competitive to state-of-the art captioning models. 
It outperforms captioning models with vision encoders, and VL-T5, 
which, like EXTRA, uses a V\&L encoder, but with object tags as linguistic inputs rather than retrieved captions. Although \extra{} does not outperform the state of the art captioning model, CaMEL, that uses a dual decoder, it outperforms the variant of CaMEL that uses the same Faster-RCNN features. \extra{} also competes with general purpose V\&L BERT models. Notice that our approach can be adapted to other V\&L encoders besides LXMERT (e.g., OSCAR, VinVL, etc.), or to more powerful decoders (e.g., as in CaMEL). Likewise, other models could benefit from retrieval-augmentation with captions.



\subsection{Ablation Studies}

We conducted a series of ablation studies in the Karpathy COCO validation split to better understand what contributes to the success of \extra{}.

\paragraph{Varying the Number of Retrieved Captions:}

We start by studying the importance of training with multiple retrieved captions, training with k=1 and k=3 captions to explore the effect of retrieving fewer captions. Table \ref{tab:ablation_k} reports the result of this experiment, showing that performance degrades when retrieving less captions. 

\begin{table}[h!]
    \centering
    \begin{tabular}{lcc}
        \toprule
        & B4 & CIDEr \\
        \midrule
              
        $k=1$ & 36.7 & 118.0 \\
        $k=3$ & 37.4 & 119.1\\
        $k=5$ & 38.3 & 121.2 \\
        \bottomrule
    \end{tabular}
    \caption{
    The effect of training and evaluating using different numbers of retrieved captions.
    Performance reported after training with cross-entropy optimization.}
    \label{tab:ablation_k}
\end{table}

\paragraph{Encoding Irrelevant Captions:}

We also studied the performance of \extra{} when it encodes textual input that is not expected to be useful. We conduct two experiments where \extra{} is trained with textual input that is either an empty caption or a 
randomly chosen caption.

\begin{itemize}
  \item \textbf{Empty Caption}: encode the image with an empty sentence: L=\{\texttt{CLS}, \texttt{SEP}\}; 
  \item \textbf{Random Caption}: encode the image with a random caption from the datastore.
\end{itemize}

Table \ref{tab:ablation_wt_retrieval} shows the result of this experiment. \extra{} outperforms both variants, further showing that the generation process is improved by encoding the image together with relevant textual context from nearest-neighbour captions. Although having an inferior performance, both models reach reasonable results compared to other models in the literature (see Table \ref{SAT}), showing that LXMERT can be used as a strong encoder for image captioning without providing relevant input image-text pairs.

\begin{table}[!h]
    \centering
    \begin{tabular}{lcc}
        \toprule
        & B4 & CIDEr \\
        \midrule
        Empty caption & 37.8 &119.1  \\
        Random caption & 37.1  & 117.7\\
        \extra{} & 38.3 & 121.2 \\
        \bottomrule
    \end{tabular}
    \caption{The effect of training and evaluating with captions that are not expected to be useful.}
    \label{tab:ablation_wt_retrieval}
\end{table}


\paragraph{Encoding Irrelevant Images:} We tested ablating the visual input (i.e.,  setting the visual features to zero). Training on ``blacked out`` input images achieves 102.1 in CIDEr, which is substantially lower than training with the actual input images, as seen in Table \ref{tab:ablation_wt_images}. This further shows that \extra{} uses the visual input, and does not just rely on the retrieved information. 

\begin{table}[!h]
    \centering
    \begin{tabular}{lcc}
        \toprule
        & B4 & CIDEr \\
        \midrule
        Blacked out image & 32.1  & 102.1\\
        \extra{} & 38.3 & 121.2 \\
        \bottomrule
    \end{tabular}
    \caption{The effect of training and evaluating with ``blacked out`` input images.}
    \label{tab:ablation_wt_images}
\end{table}

\paragraph{Changing the Retrieval System and Datastore:}

We then studied the effect of changing the retrieval system and the representations in the datastore. Recall that \extra{} relies on captions obtained by Image--Text retrieval, where the datastore contains the captions from the COCO training set, represented as vectors extracted from CLIP. 
We conducted experiments with Image--Image and Image--Text retrieval to understand which performs better:

\begin{itemize}
  \item \textbf{Image--Image Retrieval}: the datastore consists of all the images in the training data. The representation of the input image is compared against those in the datastore to find the $k$ nearest-neighbour images, and, subsequently, to obtain the $k$ captions associated to those images. Specifically, one reference caption is retrieved from each of the top-$k$ nearest-neighbour images. 
  \item \textbf{Image--Text Retrieval}: the datastore consists of all the captions associated to the images in the training data. The representation of the input image is compared against the captions to directly find the top-$k$ captions.
\end{itemize}

For Image--Image retrieval, the input image and the images in \textbf{D} are represented with \emph{Faster R-CNN} features, after global average pooling the embeddings of the 36 region-of-interest vectors. For Image--Text Retrieval, the input image and the caption vectors should already belong to a shared semantic space. We use the pretrained CLIP model because it satisfies this criteria and thus allows for direct image--text comparison. We considered two variants of CLIP based on their visual backbone: \emph{ViT} or \emph{ResNet50x4}\footnote{Regarding the comparison measure \textbf{S}, the Euclidean distance and cosine similarity were used respectively for Image-to-Image retrieval and Image-to-Text retrieval.}.
The results of this experiment are reported in Table \ref{tab:retrieval_system}.

\begin{table}[!h]
    \centering
    \begin{tabular}{lcc}
        \toprule
        & B4 & CIDEr \\
        \midrule
              
        Image--Image (Faster R-CNN) & 36.8 & 117.1 \\
        \midrule
        Image--Text (CLIP ViT) & 38.1 & 120.3\\
        Image--Text (CLIP ResNet) & 38.3 & 121.2\\
        \bottomrule
    \end{tabular}
    \caption{The effect of training and evaluating \extra{} with different retrieval systems. $k=5$ in both settings.} 
    \label{tab:retrieval_system}
\end{table}

\begin{figure}[h!]
  \includegraphics[width=1.05\linewidth]{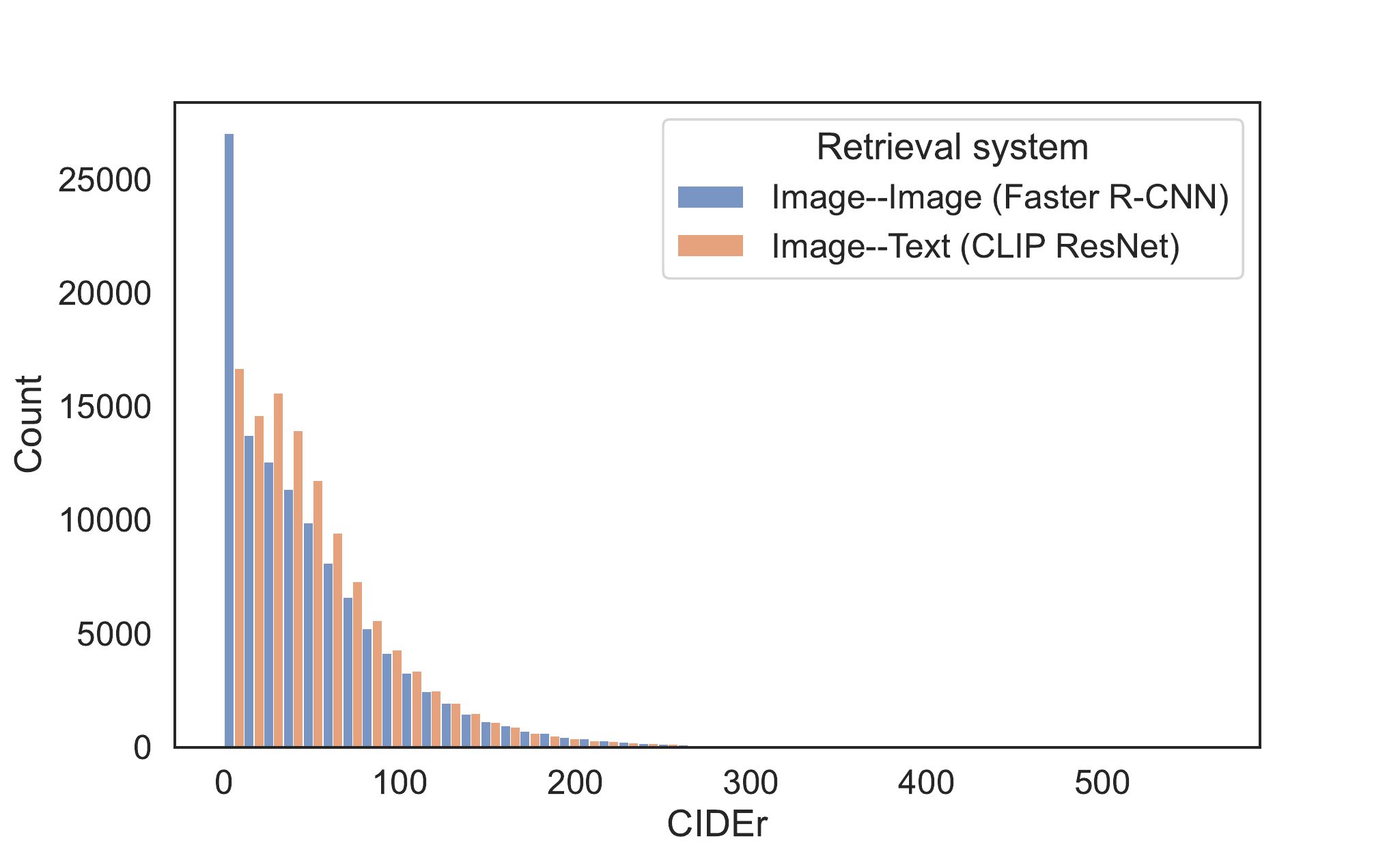}
  \caption{
  Histogram of the CIDEr scores for the nearest caption ($k=1$) retrieved with Image-Image and Image-Text retrieval. This shows the evaluations scores of using only the retrieved captions.} 
  \label{histogram}
\end{figure}

\extra{} performs worse when it uses Image--Image retrieval in comparison to retrieving captions directly with Image--Text retrieval. The best performance is obtained with the ResNet-variant of the CLIP encoder. We also assess the performance of directly using \textit{only} one of the retrieved captions, with the results shown in Figure \ref{histogram}. In this figure, we can visualize the expected CIDEr score of the first retrieved captions and observe that some of them do not sufficiently describe the image, or are mismatches, with a CIDEr of zero. We also observe that the CIDEr score can change significantly depending on the retrieval system. 
A larger number of mismatch captions are retrieved with Image-to-Image retrieval. This suggests that the retrieval system and the datastore can largely impact a retrieval-augmented image captioning model, hence they should be carefully considered.

\paragraph{Oracle Performance:}

Given that the retrieval system and datastore affect the performance of \extra{}, we also study whether \extra{} could continue to improve if it could retrieve \textit{better} captions. After training \extra{} with the $k=5$ retrieved captions, we simulate an oracle retrieval system during inference, by allowing the actual reference captions to be encoded by \extra{}. Table \ref{tab:ablation_oracle} reports on experiments in the validation data with respect to replacing one of the $k$ retrieved captions with one of the reference captions, as well as replacing all with the 5 references associated to the input images. These experiments bring a 1.8 and 8.3 point increase in CIDEr score, respectively, showing the potential for \extra{} to improve by retrieving captions that better match the input image. 

\begin{table}[t]
  \centering
  \begin{tabular}{lccccc}
    \toprule
    & B4 & CIDEr  \\
    \midrule
    $k=5$ retrieved captions & 38.3 & 121.2 \\
    $k=4$ and $1$ reference & 39.0 & 123.0 \\
    $k=0$ and $5$ references & 40.9 & 129.5  \\
    \bottomrule
  \end{tabular}
  \caption{Simulation of an oracle experiment, where \extra{} can ``retrieve'' reference captions of an image instead of retrieving all 5 captions from the datastore.}
  \label{tab:ablation_oracle}
 \end{table}
 
\section{Discussion}

\subsection{Vision First and Language Later}
\label{cross-attention-sec}

How does \extra{} use the encoded image and retrieved captions? 
We quantify this by estimating the behaviour of the cross-modal attention heads at each layer in the decoder. Specifically, we compute the average of the cross-modal attention 
across either the number of image regions or the sub-words in the encoder, at each time-step of generating a caption and across each of the 12 attention heads.

\begin{figure}[!h]
    \centering
    \includegraphics[scale=0.48]{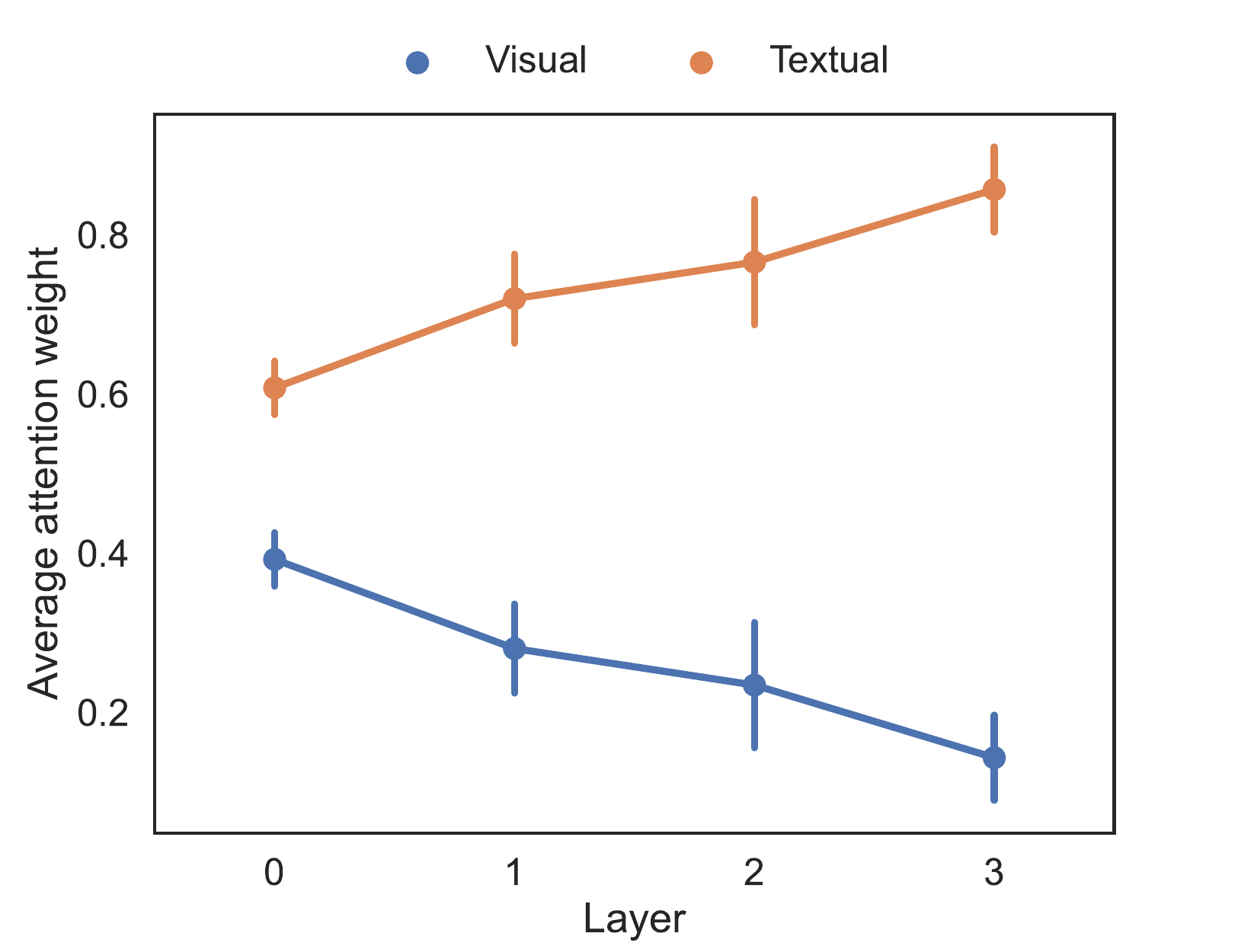}
    \caption{
    The average cross-attention from the decoder to the outputs of the encoder in respect to the visual $V$ and textual $L$ outputs. Values from COCO validation.}
    \label{fig:attention_weights}
\end{figure}

Figure \ref{fig:attention_weights} shows that across the layers, the decoder's attention shifts to the textual outputs. 
In Layer 1, the model attends both to the visual and textual representations, but the model hardly pays attention to the visual outputs by Layer 4, relying more on the textual information from the retrieved captions. This behaviour further shows that the semantics of the nearest captions can aid guiding the language generation process. We performed an identical calculation for the variants of \extra{} that encoded an empty or a random caption, finding in this case the opposite behaviour: the model learned to ignore the textual embeddings provided by the encoder (see Appendix \ref{appendix-attention}).



\subsection{Retrieve Enough Captions to Overcome Retrieval Mistakes}

We note that training with an empty set of captions was better than encoding a single $k=1$ and $k=3$ retrieved captions, observing Tables \ref{tab:ablation_k} and Table \ref{tab:ablation_wt_retrieval}. Thus, retrieval augmentation aids to improve caption quality when a sufficient number ($k=5$) is considered.
This further shows that retrieving enough captions can be crucial for success. For this, we hypothesise that retrieving more captions makes the model more robust in the presence of mismatches from certain captions, as shown for instance in the second example in Figure \ref{fig:examples}.

\subsection{Hot-swapping the Datastore}

Besides taking advantage of similar training examples, we study whether \extra{} works with external image--caption collections without needing to retrain the model. For this experiment, \extra{} was first trained and evaluated in a small dataset, and then the retrieval datastore was augmented with a larger dataset
. The considered datasets were Flickr30k and COCO, respectively. 
While Flickr30k only contains 30k images, COCO contains 113K, each paired with five sentences. Table \ref{ablation:datastore} reports the results of these experiment. \extra{} got a better performance considering a larger external dataset than just using the current training set, showing the potential for \extra{} to adapt the retrieval datastore. 

\begin{table}[!h]
  \centering
  \begin{tabular}{lccccc}
    \toprule
    Retrieval Datastore & B4 & CIDEr  \\
    \midrule
    Flickr30k & 28.8 & 59.6\\
    + COCO & 29.5 &59.9\\
    \bottomrule
  \end{tabular}
  \caption{Performance of \extra{} on the Flickr30k validation set. The model is trained on the Flickr30K dataset with the Flickr30K datastore. The datastore for inference is either the Flickr30K training set or combined with the COCO training set.}
  \label{ablation:datastore}
 \end{table}

\subsection{Qualitative Examples}

\begin{figure*}[!h]
    \centering
    \includegraphics[width=1.0\linewidth]{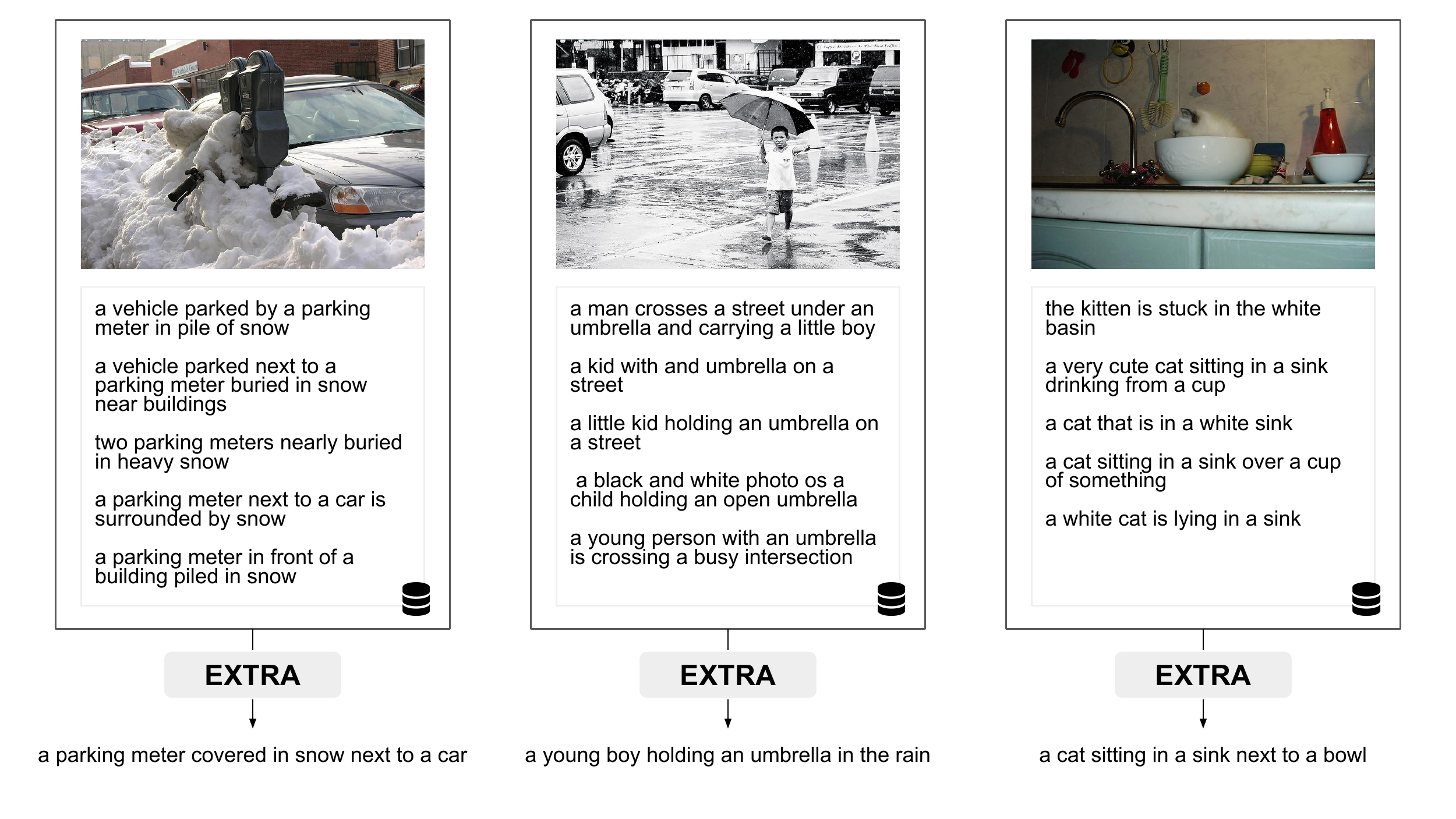}
    \vspace{-0.8cm}
    \caption{Examples of captions generated by  \extra{} conditioned on the input image and retrieved captions.}
    \label{fig:examples}
\end{figure*}

Figure \ref{fig:examples} shows examples of captions generated by \extra{}, given the input image and the $k=5$ retrieved captions. \extra{} benefits from textual evidence from nearest-neighbour captions, even though sometimes the retrieved information can be misleading, as depicted in the last example. More examples are provided in Appendix \ref{append-examples}.

\section{Related Work}  
\paragraph{Image Captioning:} The task of image captioning is usually addressed by one of these three main approaches: templates, retrieval, and encoder-decoder methods. Early approaches involved template-based methods that consisted of filling blanks of predefined captions through object detection \cite{farhadi2010every,kulkarni2013babytalk,elliott-de-vries-2015-describing}. Retrieval-based methods instead search over a dataset for the most similar image and fetch the corresponding caption \cite{hodosh2013framing,ordonez2011im2text}. Currently, the most common approach is the encoder-decoder framework \cite{xu2015show, hossain2019comprehensive}. 
The encoder typically used a pretrained CNN \cite{vinyals2016show} or a Faster R-CNN \cite{anderson2018bottom}, encoding the image into a grid of image features or object proposal image regions. The decoder was usually a LSTM with an attention mechanism \cite{xu2015show} to dynamically focus on different parts of the encoded image during the prediction of each word. 

Recently, Transformer-based models like BERT \cite{devlin2018bert} have become a more popular choice than LSTMs models, outperforming recurrent architectures in different natural language processing (NLP) tasks \cite{vaswani2017attention, qiu2020pre}. Transformers can capture long-range dependencies with self-attention layers and they can process each word of a sentence in parallel, reducing training time. After the successful application in NLP, vision Transformers like ViT \cite{dosovitskiy2020image} are also starting to become the model of choice in the field of computer vision in place of CNNs. In similar fashion, most recent captioning studies use the Transformer arquitecture \cite{herdade2019image, cornia2020meshed, liu2021cptr}, employing a vision Transformer as encoder together with an autoregressive language Transformer as decoder. Similarly to these models, this work proposes a encoder-decoder Transformer model for the task of image captioning. However, unlike them, the proposed model incorporates a pretrained V\&L BERT to exploit cross-modal representations, encoding images along with textual context. Also differently from previous work, this approach explores retrieval-augmented generation, i.e., combining neural encoder-decoder methods with traditional retrieval-based methods.  

\paragraph{V\&L BERTs:} Previous studies have proposed pretrained Vision and Language (V\&L) BERTs to learn generic cross-modal representations of images and text, that can later be used for a variety of downstream V\&L tasks \cite{bugliarello2020multimodal}. Examples include LXMERT \cite{tan2019lxmert}, VL-BERT \cite{su2019vl}, Visual BERT \cite{li2019visualbert}, OSCAR \cite{li2020oscar}, or UNITER \cite{chen2020uniter}, which were applied to VQA and other V\&L classification tasks. Given that these models are encoder-only Transformers, only few of them have been applied to generation tasks such as image captioning. In such cases, the generation is made from left to right by encoding the input image and using the textual input elements with uni-directional attention masks, 
i.e., starting with a \texttt{CLS} token with the rest of the tokens masked, then considering the \texttt{CLS} token with the predicted word (replaced by the corresponding mask token) and the remaining ones still masked, and so on \cite{li2020oscar,zhou2020unified}. 

The use of pretrained V\&L BERTs, as encoders in the standard encoder-decoder captioning framework, remains largely unexplored. The task of image captioning typically just considers single-input images, and not image-text pairs to be encoded. In our work, a pretrained V\&L encoder is used with a decoder for image captioning, by leveraging not just the images as input but also retrieved captions.

Besides pretrained V\&L encoders, pretrained V\&L encoder-decoder models have recently been proposed to tackle classification and generation tasks, such as VL-T5 \cite{cho2021unifying}. Their captioning approach is similar to the present paper, but VL-T5 uses object tags as textual inputs, whereas \extra{} is conditioned on retrieved captions. 

\paragraph{Retrieval-augmented Generation:}

The proposed approach is also similar to some studies on language generation that predict the output conditioned on retrieved examples \cite{weston2018retrieve, gu2018search, khandelwal2019generalization, lewis2020retrieval}. For instance, this work relates to \citet{weston2018retrieve}, in which a sequence-to-sequence LSTM model, for dialog generation, encodes the current input concatenated with the nearest retrieved response. Similarly, \citet{izacard2020leveraging} used an encoder-decoder Transformer conditioned on retrieved passages for open domain question answering. 
Retrieval-augmented generation is gaining traction in NLP but has only been explored for image captioning by few studies \cite{Wang_2020,fei2021memory, ramos2021retrieval, sarto2022retrieval, ramos2022smallcap}. Concurrent work proposed Transformer-based captioning models augmented with retrieval as well \cite{sarto2022retrieval, ramos2022smallcap}. However, differently from these previous studies, we encode the retrieved captions by exploiting cross-modal representations with a V\&L encoder. 

\section{Conclusions}
We propose \extra{}, a retrieval-augmented image captioning model that improves performance by exploiting cross-modal representations of the input image together with captions retrieved from a datastore. \extra{} make uses of a pretrained V\&L BERT, instead of an image-only encoder, combined with a language decoder. To generate a caption, the decoder attends to the cross-modal encoder features, containing information from image regions and also textual evidence from the retrieved captions. Image captioning is therefore addressed as language generation conditioned on vision and language inputs, instead of vision only. To evaluate this model, \extra{} was assessed against strong encoder-decoder models in the area, and ablation studies were also conducted. The experiments conducted on the COCO dataset confirmed the effectiveness of the proposed captioning approach. 

For future work, we plan to explore the utility of \extra{} in out-of-domain and in few-shot learning settings, since the retrieval component can be easily modified to include external datastores, without the need to retrain the whole model. We also plan to explore how this approach can be adapted to other powerful vision and language encoders besides LXMERT. Finally, we will explore methods that allow us to jointly train the retrieval mechanism with the full model in order to retrieve captions that are more similar to the input image.

\section*{Limitations}

Previous work has shown that generative models suffer from biases inherent to the data they are trained on \cite{weidinger2021ethical, thoppilan2022lamda}. Likewise, our EXTRA model can suffer from biases present in the COCO image captioning dataset \cite{chen2015microsoft}. Particularly, it has been shown that there is significant gender imbalance in COCO, and that captioning models can exhibit gender bias amplification (e.g., they are likely to generate the word “woman” in kitchen scenarios, and the word “man” in snowboarding scenes) \cite{hendricks2018women,zhao2017men}.

However, differently from most captioning models, EXTRA is a retrieval-augmented captioning model, and thus it has the potential to make predictions beyond the training data, by relying on information from an external datastore. Still, the datastore knowledge might also have inherent bias, as mentioned by previous studies on retrieval-augmented generation \cite{lewis2020retrieval}. In the paper, we show examples of such limitations wherein mismatched retrieved captions can bias the model towards incorrect predictions (see the results and appendix sections). 

As a way to mitigate these limitations, we recommend analyzing the corresponding nearest captions when using EXTRA, since the retrieved captions can give useful insight of the bias involved in the generation process. EXTRA can provide interpretability through textual descriptions, whereas most captioning models only provide explanations as visual attention maps.

EXTRA also has the downside of focusing on an English-centric dataset. Captioning datasets are primarily available in English, and most image captioning models are trained on COCO or other english-centric datasets. To avoid hindered research on image captioning, it is important to consider multilingual captioning datasets that contain both language-diverse captions and geographically-diverse visual concepts \cite{thapliyal2022crossmodal}.

\section*{Acknowledgements}
This research was supported by the Portuguese Recovery and Resilience Plan (RRP) through project C645008882-00000055 (Responsible.AI), and also through Funda\c{c}\~ao para a Ci\^encia e Tecnologia (FCT), namely through the Ph.D. scholarship with reference 2020.06106.BD, as well as through the INESC-ID multi-annual funding from the PIDDAC programme with reference UIDB/50021/2020. 

\bibliographystyle{acl_natbib}
\bibliography{references}

\appendix

\section{Cross-Attention}
\label{appendix-attention}

\begin{figure}[t!]
    \centering
    \includegraphics[width=1.09\linewidth]{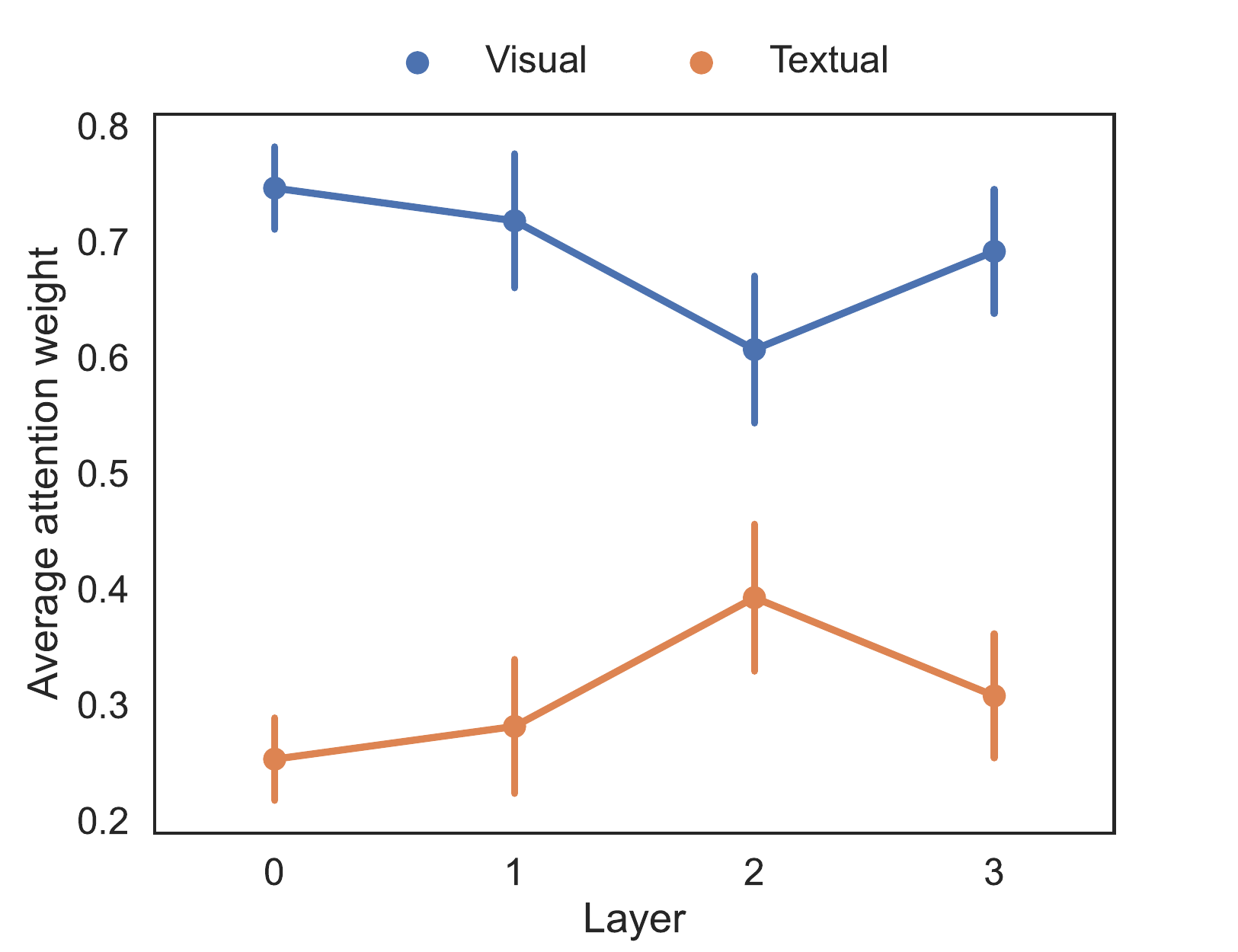}
    \caption{Cross-attention for the variant of EXTRA that
that encodes an empty caption.}
    \label{fig:cross-attention-empty}
\end{figure}

\begin{figure}[t!]
    \centering
    \includegraphics[width=1.09\linewidth]{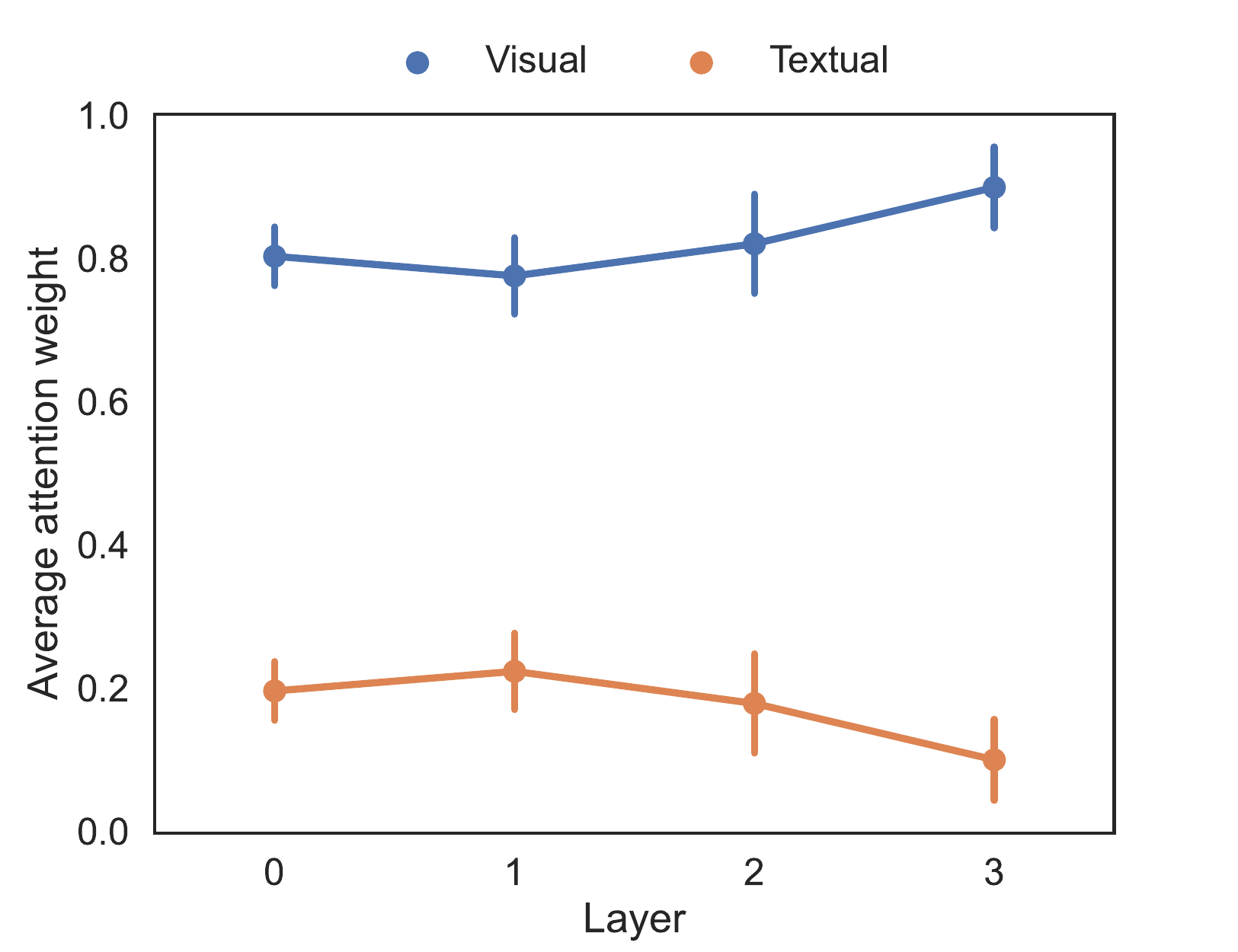}
    \caption{Cross-attention for the variant of EXTRA that
that encodes a random caption. }
    \label{fig:cross-attention-random}
\end{figure}

In Section \ref{cross-attention-sec}, we quantified how much attention EXTRA pays to the encoded image and retrieved captions. We also quantify this for the two other variants of \extra{} which encode irrelevant captions, using either an empty or a random caption. Figures \ref{fig:cross-attention-empty} and \ref{fig:cross-attention-random} show the average cross-attention weights from the decoder to the outputs of the encoder in respect to the visual \textbf{V} and textual \textbf{L} outputs, respectively for the empty and random caption encoding. Contrary to the findings presented in Section \ref{cross-attention-sec}, regarding the encoding of retrieved captions, in this scenario the two variants pay more attention to the visual outputs instead. 


For details on how we calculated the corresponding attention weights, we present the corresponding formula. Specifically, we calculated the average of the cross-modal attention $C$ across either the number of image regions or the sub-words in the encoder at each of $T$ time-step of generating a caption and across each of the $H=12$ attention heads. This calculation happens independently for each of the $L=4$ layers in the decoder:
\begin{align}
 A(C^L,V)&=\frac{1}{H}\sum_{j=1}^{H}\frac{1}{T} \sum_{t=1}^{T} \sum_{i=1}^{|V|}   \alpha_{j,t\rightarrow i}^L.\label{eqattention}\\
 A(C^L,T)&=1- A(C^L,V). \label{eqattention2}
 \end{align}

\section{More Examples}
\label{append-examples}

\begin{figure*}[h!]
    \centering
    \includegraphics[width=\linewidth]{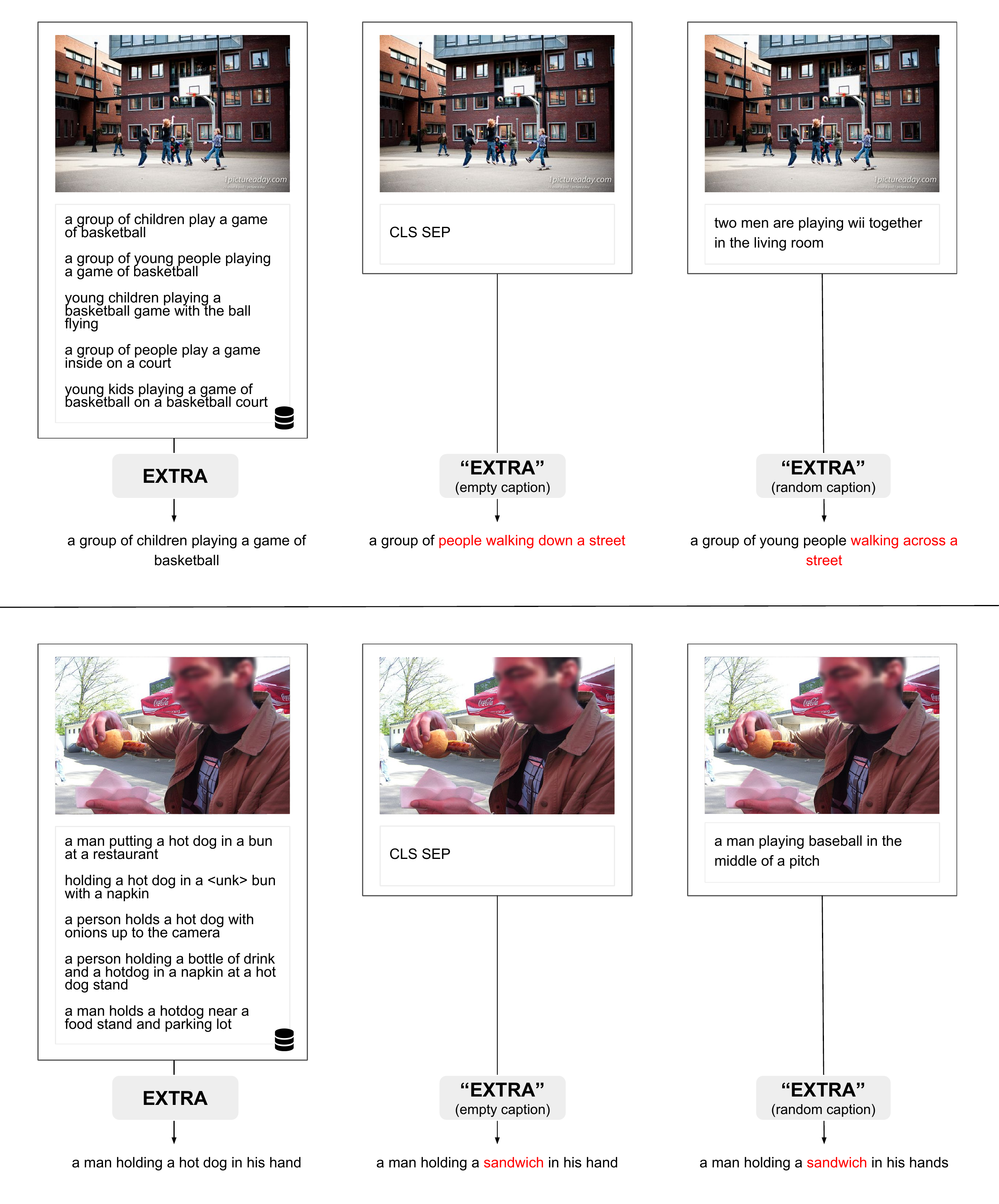}
    \caption{Examples of generated captions by \extra{} and the other two variants (empty and random caption). Better image captions are obtained from generating with retrieval augmentation.}
    \label{fig:examples_correct}
\end{figure*}

\begin{figure*}
    \centering
    \includegraphics[width=\linewidth]{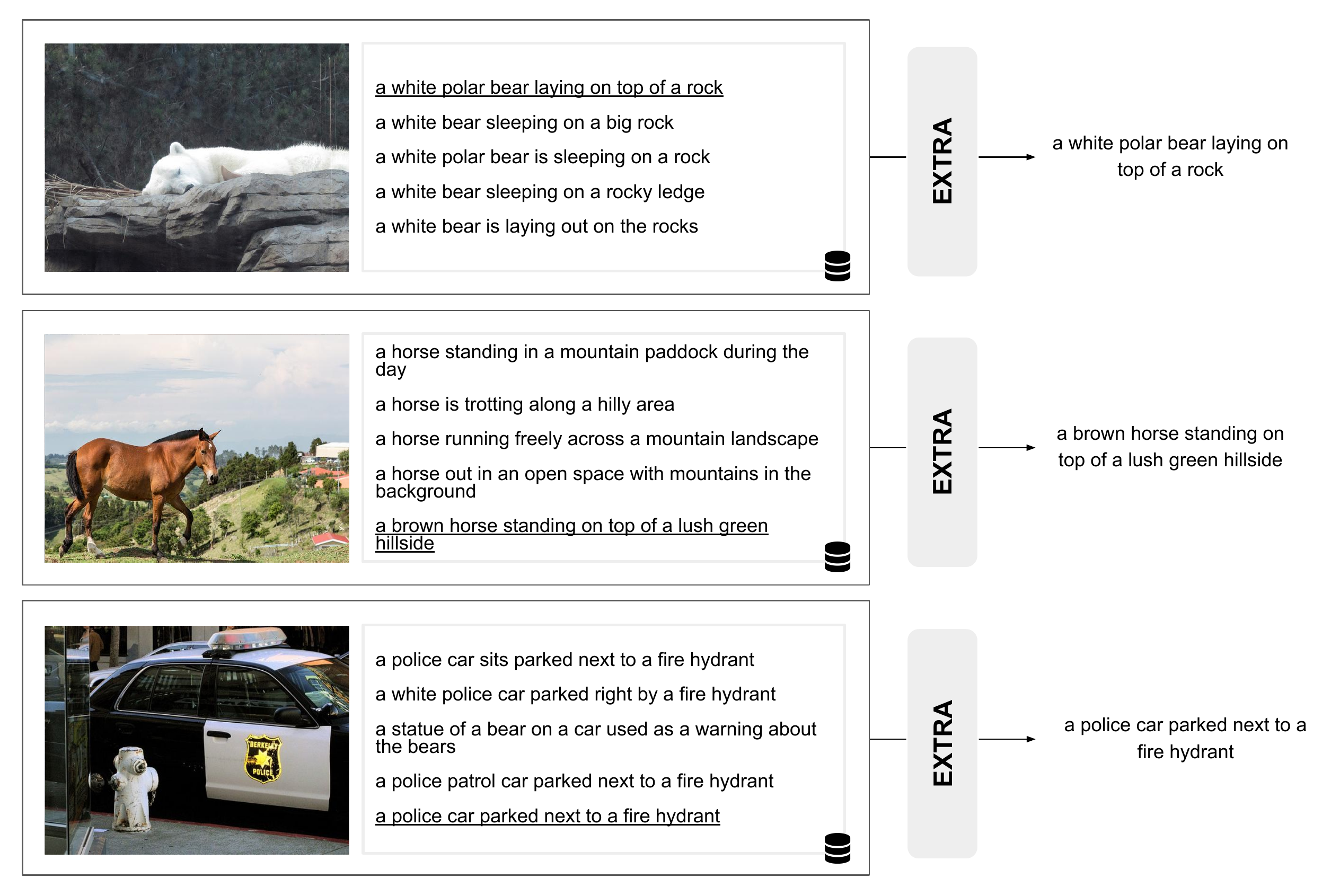}
    \caption{Examples of generated captions for which \extra{} copied from the retrieved captions.}
    \label{fig:examples_copy}
\end{figure*}

Figure \ref{fig:examples_correct} shows additional examples of the captions generated by \extra{} considering the retrieved captions, against the other two variants: encoding an empty and random caption instead. For the first image, the two variants fail to recognize that the image shows kids playing basketball (perhaps given the small size of the ball), whereas \extra{} was able to identify it by having that information in the retrieved captions. In the second image\footnote{The person was blurred for privacy concerns.}, the two variants produced the error of generating \textit{sandwich} while \extra{} correctly mentioned \textit{hot-dog}, similar to the retrieved captions. \extra{} considers the semantics from the nearest captions retrieved during generation, sometimes even copying an entire sentence, as shown in Figure \ref{fig:examples_copy}.

\begin{figure*}
    \centering
    \includegraphics[width=\linewidth]{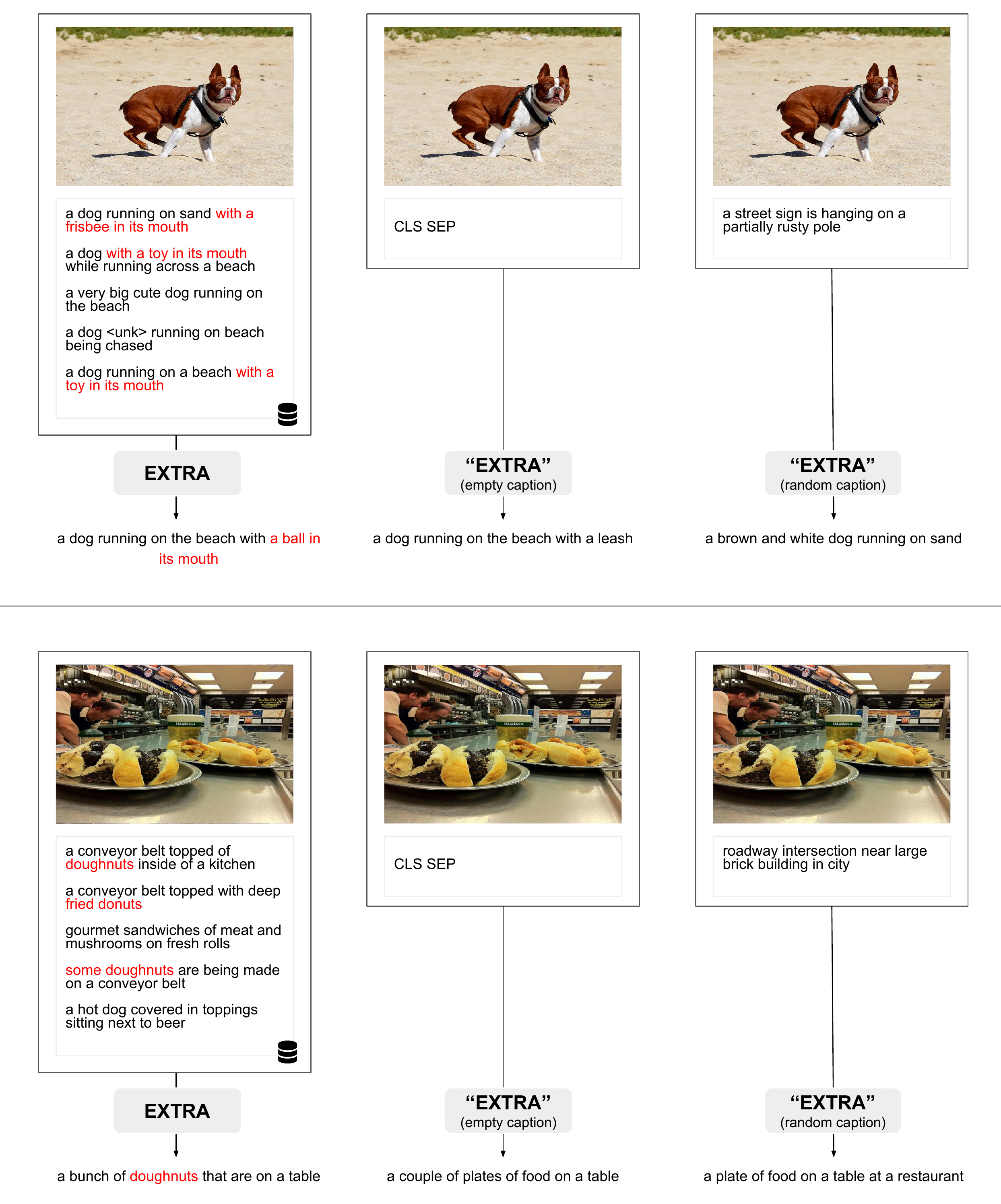}
    \caption{Qualitative results in which the retrieved
captions are not that related to the input image.}
    \label{fig:examples_incorrect}
\end{figure*}

\begin{figure*}
    \centering
    \includegraphics[width=\linewidth]{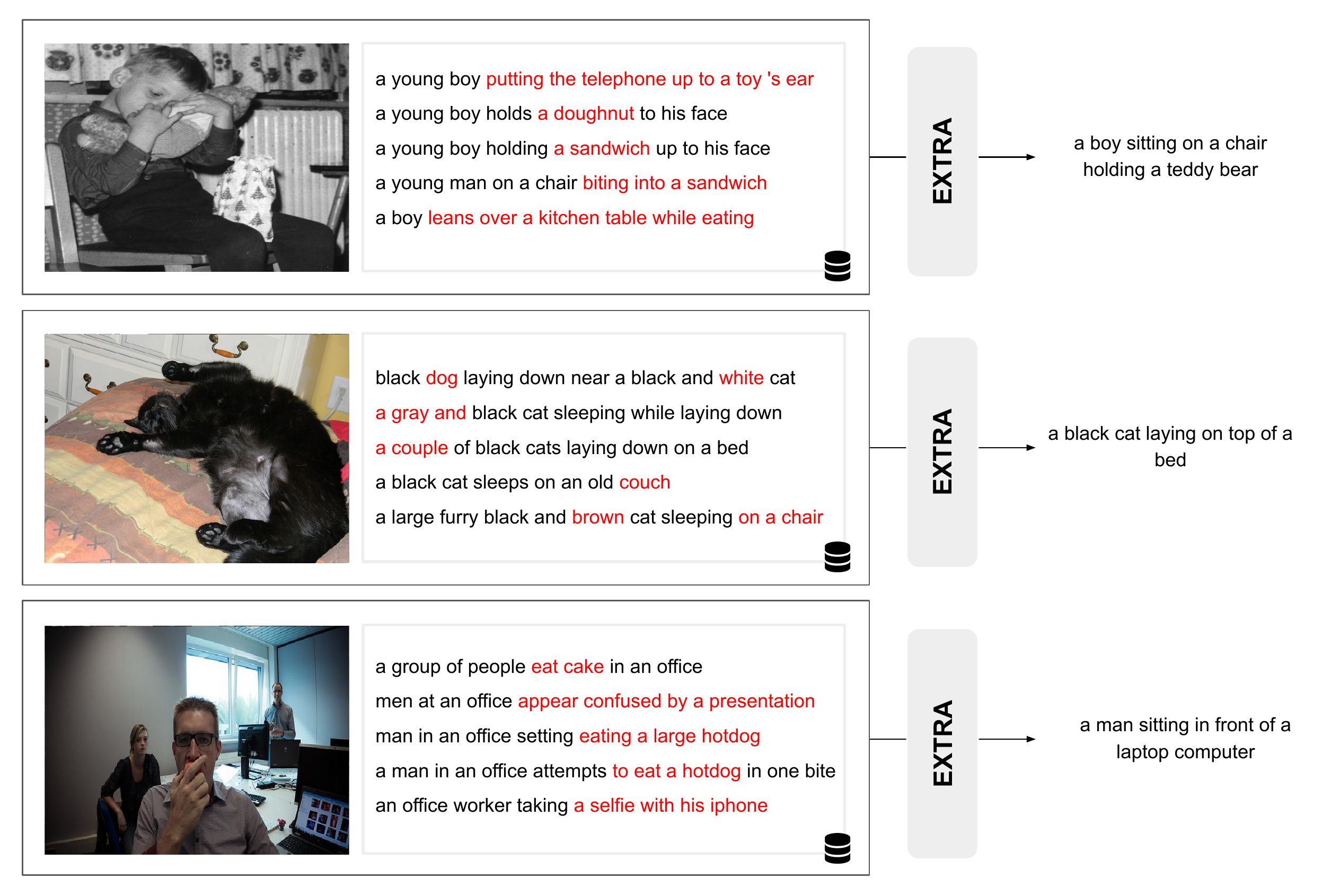}
    \caption{Examples where \extra{} is able to succeeded even with mismatches from retrieved captions.}
    \label{fig:examples_despite}
\end{figure*}

Figure \ref{fig:examples_incorrect} shows examples where the retrieved captions mislead the model. We note however that \extra{} is also able to succeed, despite the mismatch from retrieved captions, as seen in Figure~\ref{fig:examples_despite}.

\end{document}